\documentclass[conference]{IEEEtran}
\IEEEoverridecommandlockouts

\usepackage{cite}
\usepackage{amsmath,amssymb,amsfonts}
\usepackage{algorithmic}
\usepackage{graphicx}
\usepackage{textcomp}
\usepackage{xcolor}
\usepackage{booktabs}
\usepackage{multirow}
\usepackage[normalem]{ulem}
\usepackage{array}
\usepackage{hyperref} 
\def\BibTeX{{\rm B\kern-.05em{\sc i\kern-.025em b}\kern-.08em
    T\kern-.1667em\lower.7ex\hbox{E}\kern-.125emX}}

\newcolumntype{L}[1]{>{\raggedright\let\newline\\\arraybackslash\hspace{0pt}}m{#1}}
\newcolumntype{C}[1]{>{\centering\let\newline\\\arraybackslash\hspace{0pt}}m{#1}}
\newcolumntype{R}[1]{>{\raggedleft\let\newline\\\arraybackslash\hspace{0pt}}m{#1}}

\renewcommand{\paragraph}[1]{\noindent\textbf{#1}\ }

\begin{document}

\title{A Bayesian Interpretation of \\ Adaptive Low-Rank Adaptation
\thanks{This work received funding under NAST: Neural Architectures for Speech Technology, Swiss National Science Foundation grant \href{https://data.snf.ch/grants/grant/185010}{185010}.}
}

\author{\IEEEauthorblockN{Haolin Chen}
\IEEEauthorblockA{\textit{Idiap Research Institute}, Martigny, Switzerland \\
\textit{École Polytechnique Fédérale de Lausanne}, Lausanne, Switzerland \\
haolin.chen@idiap.ch}
\and
\IEEEauthorblockN{Philip N. Garner}
\IEEEauthorblockA{\textit{Idiap Research Institute} \\
Martigny, Switzerland \\
phil.garner@idiap.ch}
}

\maketitle

\begin{abstract}
Motivated by the sensitivity-based importance score of the adaptive low-rank adaptation (AdaLoRA), we utilize more theoretically supported metrics, including the signal-to-noise ratio (SNR), along with the Improved Variational Online Newton (IVON) optimizer, for adaptive parameter budget allocation. The resulting Bayesian counterpart not only has matched or surpassed the performance of using the sensitivity-based importance metric but is also a faster alternative to AdaLoRA with Adam. Our theoretical analysis reveals a significant connection between the two metrics, providing a Bayesian perspective on the efficacy of sensitivity as an importance score. Furthermore, our findings suggest that the magnitude, rather than the variance, is the primary indicator of the importance of parameters.
\end{abstract}


\section{Introduction}
In the context of the adaptation of large-scale pre-trained models, it has long been of interest to fine-tune the model in a parameter-efficient manner. Parameter-efficient fine-tuning (PEFT) techniques \cite{DBLP:journals/natmi/DingQYWYSHCCCYZWLZCLTLS23} typically optimize a small subset of the model parameters that are either original or additional ones while leaving the rest unchanged. 
The low-rank adaptation (LoRA) \cite{lora} is one of the most efficient and flexible PEFT techniques. Based on the assumption that the change of weights during fine-tuning has a low intrinsic rank, LoRA performs adaptation by optimizing the low-rank approximation of the change of the original weight matrices. Nevertheless, LoRA has limitations as it pre-defines an identical rank for all target weight matrices and therefore ignores the varying importance of weights across modules and layers. This is problematic as adding more trainable parameters to important weights contributes to better performance, however by contrast, doing so to less important weights yields marginal improvements or even inferior outcomes \cite{adalora}. 

In light of the limitation, there arises a natural question of how to allocate trainable parameters to different modules according to their importance to maximize the fine-tuning performance. To this end, a variety of techniques for LoRA has been proposed to address the problem, the most representative one of which is AdaLoRA \cite{adalora}. AdaLoRA parameterizes the delta weight mimicking the singular value decomposition (SVD) to enable dynamic adjustment of the rank: it identifies the importance of each SVD triplet in the entire model by a sensitivity-based metric and gradually prunes less important triplets during fine-tuning to reach the parameter budget. It has been demonstrated that AdaLoRA can effectively improve the model performance and parameter efficiency compared to LoRA.

Motivated by AdaLoRA, we are primarily interested in the importance scoring mechanism as it can be generically applied to PEFT for parameter selection. 
The sensitivity-based importance metric is originally based on the heuristic that the importance of parameters can be quantified by the error induced by removing them, which in turn can be approximated by the square of the gradient-weight product \cite{DBLP:journals/corr/abs-1801-05787,DBLP:conf/cvpr/MolchanovMTFK19}. 
Meanwhile, there are importance metrics with strong theoretical support, many of which originate from Bayesian neural networks (BNNs). A widely recognized metric is the signal-to-noise ratio (SNR) \cite{DBLP:conf/nips/Graves11,blundell2015weight,DBLP:conf/nips/NeklyudovMAV17}, commonly used in BNN pruning and compression. The interpretation is straightforward: a low SNR makes the neuron's output too noisy to be useful, while a high SNR indicates valuable, low-noise output. The SNR could be a drop-in replacement for the sensitivity-based importance score in AdaLoRA, allowing the pruning of SVD triplets with low SNRs during fine-tuning for dynamic rank adjustment.

The calculation of SNR requires knowledge of the variance of the parameters, typically assuming they follow a Gaussian distribution; this is closely related to variational inference (VI). VI tackles the optimization task of neural networks by approximating complex posterior distributions of the parameters; this involves selecting a simpler, parameterized distribution and minimizing the Kullback-Leibler (KL) divergence between this distribution and the true posterior. Recent advances in VI \cite{ivon} have shown not only superior performance in model calibration and predictive uncertainty compared to traditional optimizers like Adam \cite{adam}, but also high efficiency and effectiveness in large-scale networks.

In this study, we leverage Bayesian importance metrics alongside the Improved Variational Online Newton (IVON) optimizer, a recent advance in VI, to develop a Bayesian counterpart to AdaLoRA, utilizing SNR as the importance score. By comparing its performance with the sensitivity-based importance metric on the GLUE benchmark \cite{glue}, we demonstrate that the Bayesian approach not only achieves comparable or superior performance but also offers a 10\% speed-up over the original AdaLoRA with Adam. A closer examination of the underlying theory reveals a strong connection between these two metrics, providing a Bayesian interpretation of the sensitivity as an importance score. Additionally, our findings indicate that the magnitude, rather than the variance, is the primary indicator of the importance of parameters. The source code is available.\footnote{\url{https://github.com/idiap/vilora}}

\section{Adaptive Budget Allocation}
\subsection{Overview}



The techniques that enable adaptively allocating trainable parameters across different modules and layers generally fall into two categories: importance scoring-based methods and regularization-based methods. 
For importance scoring-based methods, the key is to find a proper importance metric and prune less important components accordingly. Whilst some work \cite{increlora,roselora} adopts AdaLoRA's sensitivity-based approach, other heuristic metrics, such as the magnitude of the weight \cite{dora} and the accumulated gradient \cite{rosa}, have also been explored. Among regularization-based approaches, diff pruning \cite{diff_pruning} is representative: it applies $L_0$ regularization to the delta weight (which shares the same dimensions as the pre-trained weights) and prunes it element-wise according to the magnitude. Similarly but based on LoRA, SoRA \cite{sora} introduces a gating unit in between the two LoRA matrices and applies $L_1$ regularization to the gate to zero out unimportant ranks. However, regularization-based approaches cannot guarantee to achieve target parameter budgets since they depend on unpredictable sparsity regularizations controlled by sparsity-promoting priors and threshold values, and therefore often require onerous hyperparameter tuning.

\subsection{Revisiting AdaLoRA}
AdaLoRA has the following main components.

\subsubsection{SVD-based adaptation} 
AdaLoRA parameterizes the delta weight in the form of singular value decomposition: $W = W_0 + \Delta W = W_0 + P \Lambda Q$, where $P$ and $Q$ are singular vectors and the diagonal matrix $\Lambda$ contains singular values. To avoid the intensive computational cost of SVD, a penalty $R(P, Q) = || P^\top P - I ||^2_\mathsf{F} + || Q^\top Q - I ||^2_\mathsf{F}$ is added to the loss to enforce the orthogonality of $P$ and $Q$ so that every rank is independent of each other. During adaptation, only the singular values are masked out while the singular vectors are maintained so that dropped triplets can be reactivated later.

\subsubsection{Sensitivity-based importance scoring} 
The sensitivity is defined as the magnitude of the gradient-weight product: $I(\theta) = |\theta \nabla_\theta \ell |$, where $\theta$ is a trainable parameter. The authors of AdaLoRA argue that the sensitivity itself is too variable and uncertain to be estimated due to the stochasticity of training and therefore propose to use sensitivity smoothing and uncertainty quantification: $\bar I (\theta) = \beta_1 \bar I^{t-1}(\theta) + (1 - \beta_1)I^t(\theta)$, $\bar U (\theta) = \beta_2 \bar U^{t-1}(\theta) + (1 - \beta_2)|I^t(\theta) - \bar I^t(\theta)|$,  where $\bar I^t$ is the smoothed sensitivity by exponential moving average and $\bar U^t$ is the uncertainty quantification of $I$. The final importance score is $s^t(\theta) = \bar I^t(\theta) \cdot \bar U^t(\theta)$. The authors compared its performance with the magnitude of singular values and the sensitivity without smoothing and found the proposed metric performed the best.

\subsubsection{Global budget scheduler} 
The global budget is defined as the total rank of all delta weights in the model. AdaLoRA starts from an initial budget $b^0$ that is slightly higher (usually 1.5 times) than the target budget $b^T$, warms up the training for $t_i$ steps, and gradually decreases the budget $b^t$ to reach $b^T$ following a cubic schedule. After this, the budget distribution is fixed until training finishes after $t_f$ steps.

\subsection{Bayesian Importance Scores}
In this work, we focus on theoretically supported importance metrics that originate from Bayesian neural networks (BNN). 
BNNs model weights as probability distributions, enabling the network to quantify uncertainties in its predictions. The most commonly used distribution is the Gaussian distribution, therefore the model is parameterized by two sets of parameters: the mean $\mu$ and the standard deviation  $\sigma$ (or the variance $\sigma^2$, we also refer to $\sigma$ as variance for the sake of simplicity). 

\subsubsection{\texorpdfstring{$\boldsymbol{\mathbf{SNR}(\theta)=|\mu|/\sigma}$}{SNR(theta)}}
The signal-to-noise ratio (SNR) \cite{DBLP:conf/nips/Graves11,blundell2015weight,DBLP:conf/nips/NeklyudovMAV17} is a commonly used importance metric in BNN that considers both the magnitude and the variance (also the uncertainty) of the weights. It has a simple interpretation: a low SNR results in a neuron's output being too noisy to be useful, while a high SNR signifies meaningful output with minimal noise. It has been utilized in both in-training and post-training pruning of BNNs \cite{DBLP:conf/iclr/LiMQZ24,DBLP:conf/nips/Graves11}.

\subsubsection{\texorpdfstring{$\boldsymbol{\mathbf{SNR}(|\theta|)}$}{SNR(|theta|)}} 
Li et al. \cite{DBLP:conf/iclr/LiMQZ24} argue that the random sampling of weights before each forward pass of BNN needs to be considered. Instead of using $|\mu|$ which is equal to $|\mathbb{E}_{q}\theta|$ (where $q$ is the posterior distribution of parameters), it is more appropriate to use $\mathbb{E}_{q}|\theta|$ in the SNR. The resulting metric is:
$
\operatorname{SNR}_{q}\left(\left|\theta\right|\right)=\frac{\mu\left(2 \Phi\left(\frac{\mu}{\sigma}\right)-1\right)+\frac{2 \sigma}{\sqrt{2 \pi}} \exp \left(-\frac{\mu^2}{2 \sigma^2}\right)}{\sqrt{\sigma^2+\mu^2-\left[\mu\left(2 \Phi\left(\frac{\mu}{\sigma}\right)-1\right)+\frac{2 \sigma}{\sqrt{2 \pi}} \exp \left(-\frac{\mu^2}{2 \sigma^2}\right)\right]^2}} ,
$
where $\Phi(x):=\int_{-\infty}^x \frac{1}{\sqrt{2 \pi}} \exp \left(-\frac{y^2}{2}\right) d y$ is the cumulative distribution function. It has been shown the new metric outperforms the standard SNR in training sparse BNNs \cite{DBLP:conf/iclr/LiMQZ24}.

\subsubsection{\texorpdfstring{$\boldsymbol{|\mu|}$ and $\boldsymbol{1/\sigma}$}{|mu| and 1/sigma}} 
We want to identify the key component in the SNR that reflects the importance of parameters. 
The absolute value of the mean, or the magnitude, is a straightforward metric that directly impacts the neuron's output. This metric is widely used in neural network pruning, commonly known as magnitude pruning \cite{DBLP:conf/nips/HanPTD15}. Another choice is to use the variance alone as an importance metric. The intuition is that parameters with a low variance have less uncertainty, and therefore are more important.

\section{Variational Inference}
The calculation of SNR requires approximating a Gaussian distribution over parameters, which is exactly the objective of variational inference (VI). In contrast with traditional deep learning methods that estimate parameters by minimizing the empirical risk $\ell(\boldsymbol{\theta})$ with gradient descent, variational methods estimate a posterior distribution $q(\boldsymbol \theta)$ over parameters by minimizing $\mathcal{L}(q) = \mathbb{E}_{q(\boldsymbol{\theta})}[\ell(\boldsymbol{\theta})] + \mathbb{D}_{KL}(q(\boldsymbol{\theta})||p(\boldsymbol{\theta}))$, where $p(\boldsymbol{\theta})$ is the prior. 

The optimization of $\mathcal{L}(q)$ is fundamentally different from minimizing $\ell(\boldsymbol{\theta})$ using gradient descent. For example, the expectation term requires sampling of $\boldsymbol{\theta}$ before each forward pass, and the number of parameters of $q$ is doubled for the commonly used Gaussian distribution with a diagonal covariance. Early approaches \cite{DBLP:conf/nips/Graves11,blundell2015weight} that optimize $q(\boldsymbol{\theta})$ using gradient descent have failed to scale up on modern architectures. Recent natural gradient-based methods \cite{DBLP:conf/nips/OsawaSKJETY19} have shown promising results using an Adam-like form; however, they still underperform Adam and have higher computational costs.

The Improved Variational Online Newton (IVON) \cite{ivon} is a recent VI optimizer that matches the performance of Adam at a comparable computational cost. Its key innovations include bypassing the expensive per-example gradient square computation through a reparameterization trick and incorporating several practical techniques to enhance performance.
IVON stands out as the first VI optimizer proven to be both effective and efficient for training large networks, while still delivering the benefits of VI.
In our experiments, we utilize IVON to estimate the variance of parameters, enabling the use of SNR as an importance metric.

\section{Experiments}
\subsection{Models and Datasets} 
We compare the fine-tuning performance of AdaLoRA using different importance scores on DeBERTaV3-base \cite{debertav3}. The experiments are conducted on the General Language Understanding Evaluation (GLUE) benchmark \cite{glue}, which includes four natural language inference tasks, three similarity and paraphrase tasks, and two single-sentence classification tasks.

\subsection{Implementation Details} 
We base our code on the text classification examples of the Hugging Face Transformers library \cite{hf-transformers} and the Parameter-Efficient Fine-Tuning (PEFT) library \cite{hf-peft}. For IVON, we use the official implementation\footnote{\url{https://github.com/team-approx-bayes/ivon}}. We compare the methods under two budget configurations where the target rank is set to 2 and 4 respectively, resulting in the total trainable parameters being 0.3M and 0.6M (of 86M). Full fine-tuning and LoRA applied to all modules are also added as baselines.

\subsection{Training and Evaluation} 
Our experiments are based on the official hyperparameters of AdaLoRA\footnote{\url{https://github.com/QingruZhang/AdaLoRA}} which are optimal when training with Adam. For IVON, the learning rate is set to 0.5 for MRPC and RTE and 0.4 for the rest. Same as Adam, a warm-up stage and the linear decay learning rate schedule are adopted. We found that IVON generally converges slower than Adam at the beginning of training, therefore requiring a much higher learning rate during warm-up for good results especially on small datasets. As a result, for COLA, STS-B, MRPC, and RTE, we use a higher learning rate of 2.0 in the warm-up stage and return to the normal learning rate afterwards. For evaluation, we use the best-performing model on the validation set. The results are averaged across five runs with different random seeds.

\begin{table*}[htbp]
  \centering
  \caption{Main results. The number in model names refers to the target rank. The \textbf{best} and the \uline{second best} results are marked.}
  \resizebox{\textwidth}{!}{
    \begin{tabular}{l|c|c|cccc|cccc|ccc}
    \toprule
    \multirow{4}[2]{*}{\textbf{Model}\vspace{0.5em}} & \multirow{4}[2]{*}{\textbf{Optimizer}\vspace{0.5em}} & \multirow{4}[2]{*}{\textbf{Criterion}\vspace{0.5em}} & \multicolumn{4}{c|}{\textbf{Group 1}} & \multicolumn{4}{c|}{\textbf{Group 2}} & \multirow{4}[2]{*}{\textbf{Group 1}\vspace{0.5em}} & \multirow{4}[2]{*}{\textbf{Group 2}\vspace{0.5em}} & \multirow{4}[2]{*}{\textbf{All}\vspace{0.5em}} \\
          &       &       & \textbf{MNLI} & \textbf{QQP} & \textbf{QNLI} & \textbf{SST-2} & \textbf{COLA} & \textbf{STS-B} & \textbf{MRPC} & \textbf{RTE} &       &       &  \\
          &       &       & 393k  & 364k  & 108k  & 67k   & 8.5k  & 7k    & 3.7k  & 2.5k  &       &       &  \\
          &       &       & Acc(m) & Acc   & Acc   & Acc   & Mcc   & Corr  & Acc   & Acc   &       &       &  \\
    \midrule
    $\mathrm{Full\ FT}$ & Adam  & $\mathrm{None}$  & 89.89  & 92.50  & 94.03  & 95.73  & 69.77  & 91.06  & 89.75  & 84.84  & 93.04  & 83.86  & 88.45  \\
    \midrule
    $\mathrm{LoRA}_2$ & Adam  & $r=2$ & 89.92  & \textbf{91.70} & 93.97  & 95.30  & 69.07  & 90.89  & 90.15  & 87.29  & 92.72  & 84.35  & 88.54  \\
    $\mathrm{AdaLoRA}_2$ & Adam  & $\mathrm{Sensitivity}$ & 90.40  & 91.66  & \textbf{94.49} & \uline{95.67} & \uline{70.78} & 91.47  & 90.39  & 87.00  & \uline{93.05} & 84.91  & 88.98  \\
    $\mathrm{AdaLoRA}_2$ & IVON  & $\mathrm{Sensitivity}$ & \uline{90.44} & \uline{91.69} & 94.36  & 95.55  & 69.65  & 91.89  & 90.25  & 87.94  & 93.01  & 84.93  & 88.97  \\
    $\mathrm{AdaLoRA}_2$ & IVON  & $\mathrm{SNR}(|\theta|)$ & \uline{90.44} & 91.68  & \uline{94.40} & \textbf{95.80} & 70.28  & \textbf{92.04} & 90.10  & \uline{88.16} & \textbf{93.08} & 85.15  & 89.11  \\
    $\mathrm{AdaLoRA}_2$ & IVON  & $|\mu|/\sigma$ & \textbf{90.46} & \textbf{91.70} & 94.33  & 95.62  & 70.63  & 91.90  & \textbf{90.83} & \textbf{88.38} & 93.03  & \textbf{85.43} & \textbf{89.23} \\
    $\mathrm{AdaLoRA}_2$ & IVON  & $|\mu|$ & 90.42  & \textbf{91.70} & 94.33  & 95.57  & \textbf{70.91} & \uline{91.99} & \uline{90.64} & 87.87  & 93.01  & \uline{85.35} & \uline{89.18} \\
    $\mathrm{AdaLoRA}_2$ & IVON  & $1/\sigma$ & 90.37  & 91.32  & 94.30  & 95.62  & 69.35  & 91.91  & 90.59  & \uline{88.16} & 92.90  & 85.00  & 88.95  \\
    \midrule
    $\mathrm{LoRA}_4$ & Adam  & $r=4$ & 89.68  & \textbf{92.03} & 94.13  & 95.32  & 69.58  & 90.69  & \textbf{90.25} & 87.08  & 92.79  & 84.40  & 88.59  \\
    $\mathrm{AdaLoRA}_4$ & Adam  & $\mathrm{Sensitivity}$ & 90.52  & \uline{91.91} & \textbf{94.56} & \uline{95.78} & \textbf{69.85} & 91.68  & \textbf{90.25} & 88.16  & \textbf{93.19} & 84.98  & \textbf{89.09} \\
    $\mathrm{AdaLoRA}_4$ & IVON  & $\mathrm{Sensitivity}$ & 90.54  & 91.74  & 94.45  & \textbf{95.80} & 69.41  & \textbf{92.03} & 89.90  & 88.09  & 93.13  & 84.86  & 89.00  \\
    $\mathrm{AdaLoRA}_4$ & IVON  & $\mathrm{SNR}(|\theta|)$ & \uline{90.59} & 91.78  & 94.43  & 95.67  & \uline{69.83} & 91.97  & 89.95  & \textbf{88.52} & 93.11  & \textbf{85.07} & \textbf{89.09} \\
    $\mathrm{AdaLoRA}_4$ & IVON  & $|\mu|/\sigma$ & \textbf{90.60} & 91.77  & \uline{94.49} & 95.69  & 69.32  & \uline{92.00} & 89.95  & \uline{88.30} & \uline{93.14} & 84.89  & 89.01  \\
    $\mathrm{AdaLoRA}_4$ & IVON  & $|\mu|$ & 90.56  & 91.77  & 94.40  & 95.55  & 69.72  & 91.98  & \textbf{90.25} & 88.23  & 93.07  & \uline{85.05} & \uline{89.06} \\
    $\mathrm{AdaLoRA}_4$ & IVON  & $1/\sigma$ & 90.48  & 91.22  & 94.32  & 95.62  & 69.58  & 91.93  & \uline{90.20} & 87.51  & 92.91  & 84.80  & 88.86  \\
    \bottomrule
    \end{tabular}%
   }
  \label{tab:main}%
\end{table*}%

\begin{figure}[tbp]
\centering
\includegraphics[width=1\columnwidth]{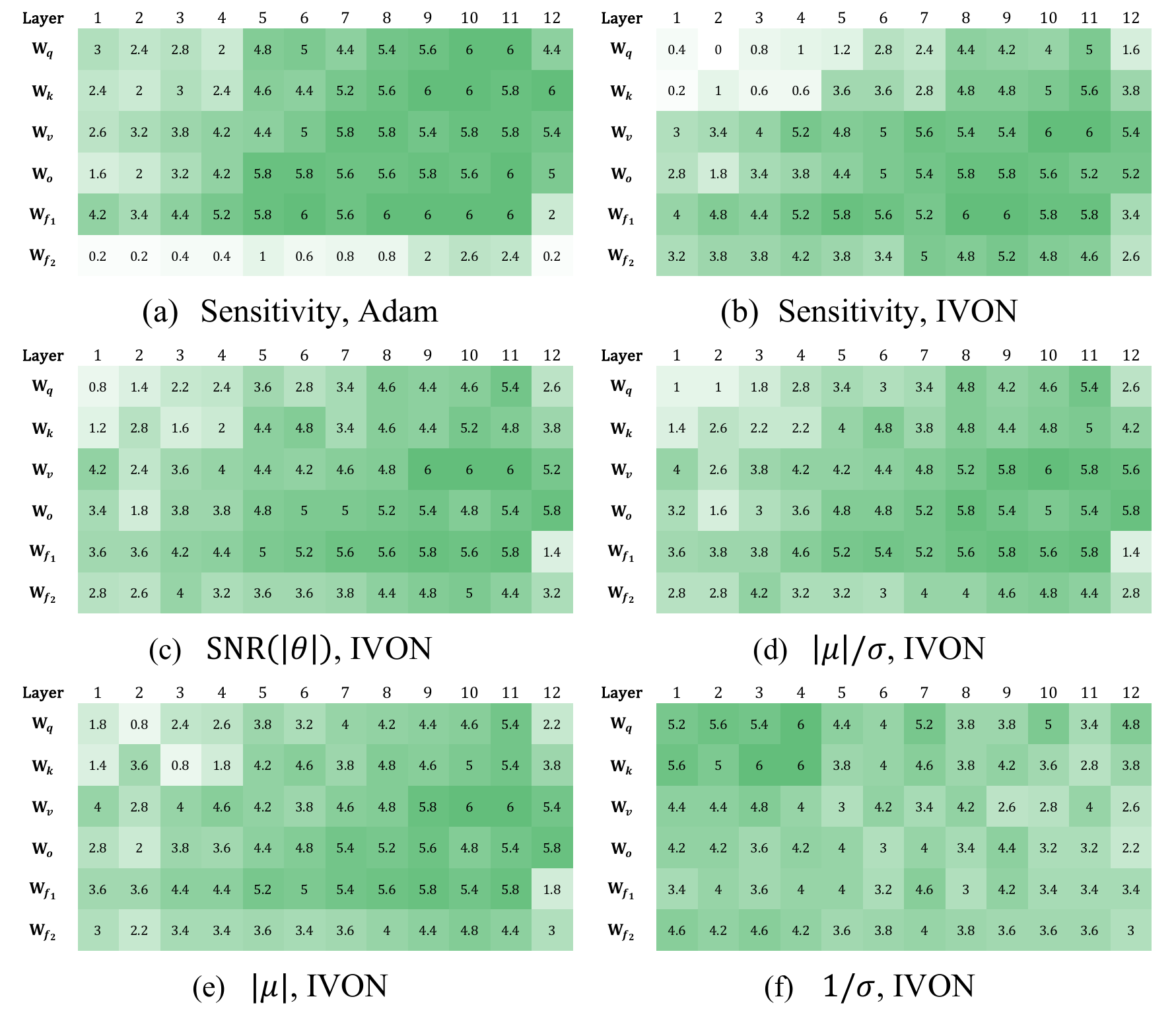}
\caption{Comparison of rank distributions after fine-tuning DeBERTaV3-base on MNLI, with deeper colors indicating higher ranks. Results are averaged across five runs with different random seeds. $W_q$, $W_k$, $W_v$, $W_o$: weights of the query, key, value, output layers of attention; $W_{f_1}$, $W_{f_2}$: weights of the feed-forward layers.}
\label{fig:rank}
\end{figure}

\section{Results and Analyses}
The main results are shown in Table \ref{tab:main}. For MNLI, the ``matched'' validation set was used for evaluation. Note that we sort the tasks according to dataset sizes and divide them into two groups since we notice that IVON needs extra tricks to ensure good results on small datasets. In general, all PEFT methods outperform full fine-tuning, and AdaLoRA outperforms LoRA. Switching the optimizer from Adam to IVON results in comparable performance, demonstrating
that IVON is capable of state-of-the-art performance in PEFT.
We further elaborate our findings from the following perspectives.

\subsection{Comparison of Importance Scores} 
Both $\mathrm{SNR}(\theta)$ and $\mathrm{SNR}(|\theta|)$ outperform the sensitivity when using IVON, and at least one of the SNR metrics outperforms or ties with the original AdaLoRA with Adam. However, there is no clear winner between the two SNR metrics. 
This could be explained by the fact that the sparsity level in the AdaLoRA case is not high (only 1/3 of the initial ranks are pruned), and that it is the SVD triplet that is pruned as a parameter group, thus the performance difference between the two metrics is not properly reflected in such a setting. 
Interestingly, the magnitude outperforms the sensitivity and one of the SNR metrics especially on small datasets. The metric was not experimented in \cite{adalora}. On the one hand, this demonstrates the effectiveness of magnitude pruning; on the other hand, this is probably because the sensitivity or the variance needs more iterations to be estimated accurately given their smoothing nature. Using the variance alone performs the worst among all metrics, however, it still outperforms LoRA with a fixed rank, indicating that the uncertainty of parameters does correlate with the importance.

\subsection{Visualizing Final Rank Distributions} 
Figure \ref{fig:rank} shows the final rank distributions of different methods after fine-tuning the model on MNLI. An obvious difference between Adam and IVON using the sensitivity can be observed comparing (a) and (b), indicating a distinction between the training dynamics of the two optimizers. The distributions of the two SNR metrics (c, d) and the magnitude (e) resemble that of the sensitivity with IVON, which corroborates with quantified results. Unlike the magnitude (e), the variance (f) shows an evenly-distributed pattern. This confirms that the magnitude plays a determining role in reflecting the importance of parameters.

\subsection{Speed} 
The variance of the parameter is inferred inherently in IVON, thus the SNR does not require the extra computation of the weight-gradient product of the sensitivity during fine-tuning. On an NVIDIA H100, using the SNR with IVON brings a 10\% speed up compared to using the sensitivity with Adam, despite the IVON itself being 2\% slower than Adam with other conditions kept the same.

\subsection{A Bayesian Interpretation of Sensitivity} 
The similarity in performance and the rank distribution between the sensitivity and the SNR suggests a close relationship between them. A closer examination of the underlying theory reveals that sensitivity is, in fact, aligned with the principles of SNR. Specifically, in IVON, the standard deviation $\boldsymbol{\sigma}$ is calculated as $\boldsymbol{\sigma} = 1/\sqrt{\lambda (\mathbf{h}+\delta)}$, where $\mathbf{h}$ is the diagonal Hessian, $\lambda$ is the effective sample size, and $\delta$ is a weight decay term. Notably, $\mathbf{h}$ can be approximated by the expected square gradient on the training data \cite{EWC}, $\mathbf{h} \approx \mathbb{E}_\mathcal{D}[(\nabla_{\boldsymbol{\theta}} \ell)^2]$, also known as the diagonal of the expected Fisher information matrix (FIM). Consequently, the inverse of the standard deviation, $1/\boldsymbol{\sigma}$, in the context of SNR, is akin to the root mean square of the gradient $\sqrt{\mathbb{E}_\mathcal{D}[(\nabla_{\boldsymbol{\theta}} \ell)^2]}$, and therefore analogous to the magnitude of the gradient $|\nabla_{\boldsymbol{\theta}} \ell|$. This implies that the sensitivity $|\theta \nabla_\theta \ell |$ has the component $|\nabla_\theta \ell|$ acting as an uncertainty measure analogous to $1/\sigma$ in SNR, thereby providing a Bayesian interpretation of the sensitivity as an importance metric. These findings resonate with the comment in \cite{DBLP:conf/cvpr/MolchanovMTFK19} that the sensitivity has connections with the FIM.
Note that both methods adopt exponential moving average smoothing to compute the global value of the corresponding metric during training. The main difference is that the smoothing is applied to the magnitude of the gradient-weight product in AdaLoRA, while the SNR is computed using the global Hessian tracked by IVON.

\section{Conclusions}
In this study, we developed a Bayesian alternative to AdaLoRA, leveraging the signal-to-noise ratio as the importance score with the IVON optimizer. By comparing the performance of different importance metrics, we demonstrated that this Bayesian approach not only matched or surpassed the performance of using the sensitivity-based importance metric on the GLUE benchmark, but was also a faster alternative to the original AdaLoRA with Adam. The theoretical analysis uncovered a significant link between these two metrics, offering a Bayesian perspective on the efficacy of the heuristic sensitivity-based metric as an importance score. Furthermore, our results suggested that the magnitude, rather than the variance, served as the key indicator of the importance of parameters.


\bibliographystyle{IEEEtran}
\bibliography{chl-ref-short}

\end{document}